# Convergent Deduction for Probabilistic Logic

Peter Haddawy
Alan M. Frisch

Department of Computer Science
University of Illinois
1304 W. Springfield Ave.
Urbana, IL 61801

**Abstract**

This paper discusses the semantics and proof theory of Nilsson's probabilistic logic, outlining both the benefits of its well-defined model theory and the drawbacks of its proof theory. Within Nilsson's semantic framework, we derive a set of inference rules which are provably sound. The resulting proof system, in contrast to Nilsson's approach, has the important feature of *convergence* - that is, the inference process proceeds by computing increasingly narrow probability intervals which converge from above and below on the smallest entailed probability interval. Thus the procedure can be stopped at any time to yield partial information concerning the smallest entailed interval.

## 1. Introduction

Expert systems often need to deal with incomplete and uncertain information which stems from a variety of sources. One source of this imprecision is the experts themselves who, in many cases, can provide only heuristic rules describing their decision processes. In addition, an expert system must draw inferences from imprecise observations, such as when these observations are distorted by noise in the environment. Lastly, incomplete information can result from lack of sufficient time to gather all the facts relevant to a particular decision.

The field of Artificial Intelligence has taken several approaches toward developing systems capable of reasoning under uncertainty. Unfortunately, most of the solutions proposed to date either are ad hoc or are based on unreasonable assumptions concerning the structure of the domain. For example, the certainty factor method of the Mycin system [Shortliffe, 1976] has no clear theoretical basis. The Bayesian approach taken in the Prospector system [Duda, et. al., 1979] deviates from proper Bayesian decision theory in order to gain efficiency. Dempster-Shafer theory [Dempster, 1967, 1968; Shafer, 1976], which has gained much popularity among AI researchers in recent years (e.g., Ginsberg [1984] and Garvey, et.al. [1981]), is plagued by a number of problems, both theoretical and practical. In particular, Dempster's combination rule does not handle conflicting evidence properly [Zadeh, 1979; Rosenkrantz, 1981].[1] Furthermore, the combination rule assumes conditional independence among multiple bodies of evidence supporting a single hypothesis, an assumption that is unwarranted in many domains.

---

[1] For example, if a frame of discernment contains three mutually exclusive alternatives and two bodies of evidence each strongly support a different one of the alternatives, then the combination rule will place all the combined support on the third unsupported alternative.

Clearly a theoretical framework is needed which will facilitate analysis and comparison of existing uncertainty calculi and will allow development of uncertainty calculi with well defined properties. This requires the definition of a language with a formally defined semantics. The semantics should be framed in terms of probability theory in order to allow analysis of the rationality of any calculi developed within the language. Nilsson's [1986] probabilistic logic is one such language.

In probabilistic logic each model assigns a probability to every first-order sentence. Probabilistic logic is appealing in many ways. Its well-defined model theory clarifies the meaning of the probability attached to a sentence. The model theory also yields clear definitions of consistency, validity, and entailment independent of any system of inference. Thus, the model theory decisively settles any questions about whether one inference rule or another is justified.

Accompanying his presentation of probabilistic logic, Nilsson also presents a procedure for computing *probabilistic entailment*. Given a set of first-order sentences and their associated probabilities, this procedure computes a range of probabilities within which the probability of some given target sentence must lie. The procedure operates by first finding all consistent assignments of truth values to the given sentences and then using these assignments to set up a system of linear equations that must be solved.

Nilsson's proposal for computing probabilistic entailment has a number of drawbacks however. The foremost problem with his approach is that it requires determining all consistent truth value assignments for a set of sentences, a problem which is NP-complete for propositional logic and undecidable for first-order logic. Hence, in the first-order case the method may not proceed even as far as setting up the system of linear equations, in which case the method yields no information about the probability of the target sentence.

Another drawback of Nilsson's procedure is that it does not produce a proof in the usual sense. A traditional deductive system based on inference rules yields a proof that explains the line of reasoning used and justifies the conclusion. Such proofs lend themselves to machine analysis and manipulation, such as creation of generalized macro operators.

This paper presents a generalization of Nilsson's logic and reports on progress we have made towards developing a *convergent* proof system based on inference rules. The property of convergence means that the inference process proceeds by computing increasingly narrow probability intervals that converge from above and below on the smallest entailed probability interval. Thus the procedure can be stopped at any time to yield partial information concerning the probability range of the target sentence. Consequently, with a convergent deductive system one can make a tradeoff between precision and computation time.

In particular, this paper presents the derivation of several sound inference rules for probabilistic logic and comments on the difficulty of constructing a complete set of inference rules. It then presents an example that demonstrates the convergent nature of the proof rules and finally suggests several extensions to the logic. However, before turning to these topics Nilsson's proposals are first described from our own vantage point. For simplicity the discussion is limited to propositional probabilistic logic.

## 2. Nilsson's Proposal

The language of probabilistic propositional logic is that of ordinary propositional logic. However, in probabilistic logic, the semantic value of a sentence is a probability. Probabilistic logic reduces to ordinary logic if the semantic value of every sentence is either 0 (False) or 1 (True). In order to assign probabilities to sentences, a model is defined to contain a non-empty finite[2] set of possible worlds and a probability distribution over the worlds.[3] The probability of a world is the probability that it is the actual world. The semantic value of a sentence is simply the sum of the probabilities of the worlds that satisfy the sentence.

Consider now the notion of probabilistic entailment, the analogue of logical entailment for probabilistic logic. Given a set of sentences, $B$, and their probabilities, we wish to derive the probability of a given target sentence $S$. Nilsson calls this the probabilistic entailment of $S$ from $B$. To state this explicitly in terms of the model theory, consider all models that assign the sentences of $B$ their stipulated probabilities and ask what probabilities these models assign to $S$. In general, these models do not assign to $S$ a unique probability, but rather a range of probabilities. This range is the probabilistic entailment of $S$ from $B$.

Nilsson proposes a procedure for computing probabilistic entailment. The procedure first finds every possible assignment of truth and falsity that a world can give to the sentences of $B \cup \{S\}$. These assignments are then used to set up a system of linear equations that express the constraints between the probabilities of the sentences in $B \cup \{S\}$ and the probability distribution over the possible worlds. Solving these equations for the probability of $S$ in general results in an inequality of the form $x \leq \text{prob}(S) \leq y$.

In propositional probabilistic logic, this method provides a decision procedure for finding the probabilistic entailment of $S$ from $B$. However, for a first-order probabilistic logic this method does not provide even a semi-decision procedure. The difficulty is that if $B \cup \{S\}$ is consistent then the set of its possible truth assignments is not enumerable. Therefore, Nilsson's method may not even proceed as far as setting up the system of linear equations. In this case, the method yields no information about the probability of $S$.

Nilsson's procedure is all or none; it either computes the probabilistic entailment of $S$ from $B$ or it yields no information about it. As such, it fails to exploit the capacity of probabilistic logic (and, in fact, any multi-valued logic) to express intermediate results. Before the ultimate value, or value range, of the target sentence is computed, it is possible to have an intermediate result stating that some truth values have been eliminated. Notice that a two-valued logic does not have this capacity; once one value is eliminated, the value of the target sentence is fully determined.

## 3. A Modal Logic of Probability

We generalize Nilsson's probabilistic logic to allow sentences to explicitly state the interval in which the probability of a proposition lies. This is necessary

---

[2] By limiting our attention to a propositional logic we need only consider finite sets of worlds, thereby avoiding the difficulty of dealing with a probability distribution over an infinite set. Nilsson addresses these problems.

[3] For a philosophical discussion of the use of modal logic in formalizing the semantics of pro-

for the deductive system we wish to develop since, as Nilsson has shown, even if we start with point probabilities for the axioms we will in general infer indeterminate probabilities for any given statement that follows from the axioms. We would like to use these probability intervals to infer further statements. Thus we simply allow indeterminate probabilities to be assigned to axioms. This is equivalent to Grosof's [1986] formulation of Type-1-ui axiom sets. In this framework we now define the modal language of probability, $L_{\text{PL}}$, with its notions of model, satisfaction, and entailment.

The language that we use explicitly states the range of probabilities that are associated with each sentence of propositional logic. Accordingly, sentences of $L_{PL}$ are of the form $\mathbf{P}\Phi \in [x,y]$, where $\mathbf{P}$ is the modal operator, $\Phi$ is any sentence of propositional logic, and $[x,y]$ names a subinterval of the unit interval. This paper continues to use $\Phi$ and $[x,y]$ in this way and also uses $I$ as the name of a probability interval. Semantically, the value assigned to the expression $\mathbf{P}\Phi$ in the model $M$, written $\|\mathbf{P}\Phi\|^M$, is the probability assigned to $\Phi$ by $M$ - that is, the sum of the probabilities of the worlds that satisfy $\Phi$. Thus the semantic values of the expressions $\mathbf{P}\Phi$ are governed by the axioms of probability. The semantic value of $\mathbf{P}\Phi \in [x,y]$ is defined as one would expect:

$\|\mathbf{P}\Phi \in [x,y]\|^M = $ True if $x \leq \|\mathbf{P}\Phi\|^M \leq y$
$\qquad\qquad\qquad\quad = $ False otherwise

Thus we have a two-valued logic that explicitly talks about probabilities. Consequently, validity, consistency, and entailment are defined as in any other two-valued logic. For example, we have:

> A set of $L_{PL}$ sentences $B$ entails the $L_{PL}$ sentence $S$ (written $B \models_p S$ iff every probabilistic model which satisfies $B$ also satisfies $S$.

In general, there will be many intervals $[x,y]$ such that $B \models_p \mathbf{P}\ \Phi \in [x,y]$. From among these intervals, a probabilistic entailment procedure is concerned with finding the one that is the least, as ordered by the subset relation. We call this the *least entailed interval of* $\Phi$ *from* $B$. The existence of such an interval is guaranteed by the following theorem.

**Theorem:** For any set of $L_{PL}$ sentences $B$ and any propositional sentence $\Phi$, there is a least interval, $I$ such that $B \models_p \mathbf{P}\ \Phi \in I$.

**Proof:** Consider $Q$, the set of all probabilities assigned to $\Phi$ by the models that satisfy $B$. We show that there is a least interval containing $Q$. If $Q$ is empty, then the empty interval is the least containing interval. Otherwise $Q$ is a bounded (bounded by 1) nonempty subset of the reals, and therefore has a least upper bound, y, in the reals. (See theorem 5RB of Enderton [1977].) Similarly, $Q$ has a greatest lower bound, y, in the reals. Therefore, $[x,y]$ is the least interval containing $Q$.

Before turning to our deductive system, we note several semantic properties, which are immediate consequences of the semantic definitions of this section.

---

bability see [Carnap & Jeffrey, 1971].

**Semantic Properties**

Let $M$ be an arbitrary probabilistic model and let $\alpha$ and $\beta$ be arbitrary propositional sentences. Then:

1) if $\alpha$ is valid then $\|\mathbf{P}\alpha\|^M = 1$
2) if $\alpha$ is inconsistent then $\|\mathbf{P}\alpha\|^M = 0$
3) if $\alpha$ is logically equivalent to $\beta$ then $\|\mathbf{P}\alpha\|^M = \|\mathbf{P}\beta\|^M$
4) $\|\mathbf{P}\alpha\|^M = 1 - \|\mathbf{P}\neg\alpha\|^M$
5) $\|\mathbf{P}\alpha\vee\beta\|^M = \|\mathbf{P}\alpha\|^M + \|\mathbf{P}\beta\|^M - \|\mathbf{P}\alpha\&\beta\|^M$
6) if $I_1 \subseteq I_2$ then $\mathbf{P}\alpha \in I_1 \models_p \mathbf{P}\alpha \in I_2$.
7) $\mathbf{P}\alpha \in [0,1]$ is valid.

## 4. Convergent Deduction

In computing, or attempting to compute, the probabilistic entailment of a certain sentence we shall be concerned with deduction procedures that throughout their execution maintain an estimate of the the least entailed interval. This estimate will be referred to as the *current derived interval*. A procedure that operates in this manner can, at any time, be asked to report its current derived interval and thus can provide information about the least entailed interval before it computes exactly what that interval is.

We can now describe the notions of soundness, completeness, convergence, and monotonicity for such a deductive procedure. If a procedure is such that its current derived interval always contains the least entailed interval, then the procedure is *sound*. The deductive procedure is called *convergent* if its current derived interval converges in the limit. That is, there is some point of time after which the procedure's current derived interval never changes. The procedure is said to be *monotonic* if its current derived interval only ever changes to a subset of its value. Monotonicity guarantees that that the deduction procedure will never derive a worse answer by running longer. Finally, a convergent deductive system is *complete* if it converges to the least entailed interval.

Notice that nothing here says that we must know when a convergent procedure has converged. Indeed, in the first-order case such knowledge is precluded by undecidability.

Given a set of sound inference rules for $L_{PL}$ it is simple to come up with a sound monotonic deductive procedure for computing the probabilistic entailment of $\Phi$ from $B$. The procedure begins with $[0,1]$ as its current derived interval and proceeds by enumerating all possible proofs from $B$ that can be constructed with the given inference rules. Suppose that at some point the procedure has a current derived interval of $I_1$ and that it generates a proof of $\mathbf{P}\Phi \in I_2$. What should the procedure do at this point? In some probabilistic calculi this would be considered as two pieces of possibly conflicting evidence that must be reconciled. In a non-monotonic logic, the appropriate action might be to retract one of the statements. However, in the present system it is clear that, since both of these intervals are derived by sound rules of inference, the probability of $\Phi$ must lie within *both* intervals. Accordingly, the current derived interval should be updated to $I_1 \cap I_2$. Notice that if the current derived interval is ever the empty interval–as in the previous example if $I_1$ and $I_2$ were disjoint–then the deductive procedure has concluded that $B$ is inconsistent.

## 5. The Inference Rules

This section presents five sound inference rules for $L_{PL}$. The inference rules are listed below, followed by derivations of the first two rules. The remaining derivations are similar and are omitted for lack of space. Note that rules 1 through 3 are identical to those presented by [Garvey, et.al., 1981]. A derivation of rule 3 can be found in [Dubois & Prade, 1985].

1) The negation rule:
   $\mathbf{P}\ A \in [x,y] \vdash \mathbf{P}\ \neg A \in [1-y, 1-x]$

2) The conjunction rule:
   $\{(\mathbf{P}(A) \in [x,y], \mathbf{P}(B) \in [u,v])\} \vdash \mathbf{P}(A\&B) \in [\max(0, x+u-1), \min(y,v)]$

3) The implication rule:
   $\{\mathbf{P}A \in [x,y], \mathbf{P}(A \rightarrow B) \in [u,v]\} \vdash \mathbf{P}B \in [\max(0, x+u-1), v]$

4) Combining the rules for conjunction and implication yields the following rule for propagation across Horn-clauses:
   $\{\mathbf{P}\ (A_1 \& \cdots \& A_n \rightarrow B) \in [x,y],\ \mathbf{P}\ A_i \in [u_i, v_i]\}$
   $\vdash \mathbf{P}\ B \in [\max(0, x + \sum u_i - n), y]$

5) The multiple derivation rule:
   $\{\mathbf{P}\Phi \in [x,y], \mathbf{P}\Phi \in [u,v]\} \vdash \mathbf{P}\Phi \in [\max(x,u), \min(y,v)]$

### Derivation of the Negation Rule

Consider an arbitrary model $M$. By semantic rule 4,

$$\|\mathbf{P}\ \neg A\|^M = 1 - \|\mathbf{P}\ A\|^M$$

Suppose that $x \leq \|\mathbf{P}\ A\|^M \leq y$, then

$$1-y \leq \|\mathbf{P}\ \neg A\|^M \leq 1-x$$

Therefore, from $\mathbf{P}\ A \in [x,y]$ one can derive $\mathbf{P}\ \neg A \in [1-y, 1-x]$ and vice versa.

### Derivation of the Conjunction Rule

To derive the rule for conjunction, suppose we wish to find $[w,z]$, the least entailed interval of A&B from $\mathbf{P}A \in [x,y]$ and $\mathbf{P}B \in [u,v]$. The value $w$ can be obtained by considering two cases: i) $x+u \leq 1$ and ii) $x+u > 1$. In the first case, A&B can be inconsistent; thus $w$ is just 0. The second case is more complicated. By semantic property 5, $\|\mathbf{P}\ A\|^M + \|\mathbf{P}\ B\|^M = \|\mathbf{P}\ A\&B\|^M + \|\mathbf{P}\ AvB\|^M$
Combining this with the given probability ranges,

$$\|\mathbf{P}\ A\&B\|^M + \|\mathbf{P}\ AvB\|^M \geq x+u > 1$$
$$\|\mathbf{P}\ A\&B\|^M \geq x+u - \|\mathbf{P}\ AvB\|^M > 1 - \|\mathbf{P}\ AvB\|^M$$

The least upper bound on $\|\mathbf{P}\ AvB\|^M$ under the assumption for this case is just 1. So

$$\|\mathbf{P}\ A\&B\|^M \geq x+u-1 > 0$$

Combining the two lower bounds yields

$$\|\mathbf{P}\ A\&B\|^M \geq \max(0, x+u-1)$$

To derive the upper bound first note that by the given probability ranges

$\|\mathbf{P}\ A\|^M + \|\mathbf{P}\ B\|^M \leq y+v$

This can be combined with semantic property 5 to give

$\|\mathbf{P}\ A\&B\|^M \leq y+v - \|\mathbf{P}\ AvB\|^M$

The sum on the right is at a maximum when $\|\mathbf{P}\ AvB\|^M$ is at a minimum, which occurs when either A→B or B→A. Thus the lower bound on $\|\mathbf{P}\ AvB\|^M$ is max(y,v). Substituting this into the above inequality and simplifying yields

$\|\mathbf{P}\ A\&B\|^M \leq \min(y,v)$

Combining the upper and lower bounds we finally have that

$\{(\mathbf{P}(A)\in[x,y],\ \mathbf{P}(B)\in[u,v])\}\ \vdash\ \mathbf{P}(A\&B)\in[\max(0,x+u-1),\ \min(y,v)]$.

The reader will note that nothing has been proven concerning the completeness of the inference rules. Preliminary work has shown that there are special cases where these rules are provably complete. However, the set of rules as it stands is in fact provably incomplete for propositional logic in general. To be complete, any set of inference rules must embody the axioms of probability. One of these axioms is the following:

P(AvB) = P(A) + P(B) if A and B are contradictory.

No application of the presented set of proof rules is capable of deriving this fact. This axiom requires knowledge of the consistency of A and B. It remains an open question whether our current set of sound inference rules can be augmented in such a way that it embodies all the axioms of probability and thus yields a complete set of convergent inference rules for probabilistic logic.

## 6. Example

We now present a simple example of the deductive system in action. Suppose we have the following statements:

$\mathbf{P}(A\&B \rightarrow C) \in [.8\ .9]$
$\mathbf{P}A \in [.7\ .8]$
$\mathbf{P}B \in [.8\ 1]$
$\mathbf{P}(D\rightarrow C) \in [.7\ .8]$
$\mathbf{P}D \in [.5\ .7]$

from which we wish to find the probabilistic entailment of C. The system starts with [0,1] as its current derived interval. Supposing that at some point it uses the Horn clause rule to derive $\mathbf{P}C \in [.3\ .9]$ from the first three statements, the procedure updates its current derived interval to [.3 .9]. Supposing that it then uses the Horn clause rule to derive $\mathbf{P}C \in [.2\ .8]$ from the last two statements, it can then use the rule for multiple derivations to update its current derived interval to [.3 .8], which is, in fact, the least entailed interval. Notice how the current derived interval monotonically converged on the least entailed interval.

## 7. Summary and Future Research

We have identified and discussed the important notion of convergence in probabilistic inference. Convergent inference methods can return useful partial information even before a complete proof is found. This is particularly useful when dealing with first-order probabilistic logic. We showed that Nilsson's method for computing probabilistic entailment is not convergent. We then described a modal logic of probability which is a slight generalization of Nilsson's logic and formulated a set of convergent inference rules which are sound with respect to this model theory. Additionally, these rules embody no assumptions concerning the probability distribution over possible worlds. They are similar to those used in the INFERNO system [Quinlan, 1983] and require little more than a Prolog type interpreter for their implementation.

This paper thus far has dealt exclusively with marginal probabilities. Although it is possible to express conditional probabilities in terms of marginals–i.e. $P(A|B) = P(A\&B)/P(B)$– specification of the conditional probability requires less information than specification of the two necessary marginal probabilities. Additionally, conditional probabilities are a more general representation in the sense that the probability of any statement can be considered to be conditioned on the tautological proposition. Given their utility and generality, it seems important to have an explicit representation for conditional probabilities. Fortunately, this amounts to a simple extension to our logic. Since the probability of A given B is simply the probability of A within those worlds which satisfy B, the semantic value of the expression $\mathbf{P}(A|B)$ can be defined to be the sum of the probabilities of the worlds that satisfy A&B divided by the sum of the probabilities of the worlds that satisfy B. A suitable method for using conditional probabilities to update on new evidence remains to be formulated although some form of Bayesian inference is the obvious choice.

The real power of probabilistic logic and convergent deduction attains when quantification is introduced. Although this paper has not addressed the problem of quantification, it presents no theoretical difficulties within the semantic framework. The problem of quantification inside and outside the scope of a modal operator is a well understood problem in logic. Earlier attempts at dealing with quantification such as [Ginsberg, 1984] have left vague the meaning of statements such as "the probability that all birds fly is 0.9." We, on the other hand, by deferring to the model theory, can provide a precise definition for the meaning of quantified statements. For example, we can distinguish between the two statements $\mathbf{P} \ \forall x A(x) \in I$ and $\forall x \mathbf{P} \ A(x) \in I$. While the first statement is true iff the sum of probabilities of the worlds satisfying $\forall x A(x)$ is in the interval $I$, the second is true iff for each x, the sum of the probabilities of the worlds satisfying $A(x)$ is in the interval $I$. The model theory shows that the two statements have completely different meanings. A full theory of quantification and of multiple nested modal operators is a topic for further research.

### Acknowledgements

We would like to thank Patrick Maher for his insightful comments on a draft of this paper and Lisa Wolf whose editorial comments greatly improved the presentation.